\documentclass[runningheads, anonymous]{llncs}
\usepackage[T1]{fontenc}
\usepackage{graphicx}
\usepackage{cite}
\usepackage{adjustbox}
\usepackage{tikz,pgfplots,pgfplotstable}
\usepackage{amsmath}
\usepackage{amssymb}
\usepackage{multirow}
\usepackage{pifont}
\usepackage{xcolor, soul}
\usepackage{subcaption}

\newcommand{\figref}[1]{Fig.~\ref{#1}}
\newcommand{\secref}[1]{Section~\ref{#1}}
\newcommand{\tabref}[1]{Table~\ref{#1}}

\usepackage{hyperref}
\hypersetup{
    colorlinks=true,
    linkcolor=blue,
    filecolor=blue,      
    urlcolor=blue,
    citecolor=blue,
    pdftitle={CHAD},
    }

\usepackage{color}

\begin{document}
\title{CHAD: Charlotte Anomaly Dataset}

\author{Armin Danesh~Pazho \and
Ghazal Alinezhad~Noghre \and
Babak Rahimi~Ardabili \and
Christopher Neff \and
Hamed Tabkhi
}

\authorrunning{Danesh~Pazho et al.}

\institute{University of North Carolina at Charlotte, Charlotte NC 28223, USA \email{adaneshp@uncc.edu, galinezh@uncc.edu, brahimia@uncc.edu, cneff1@uncc.edu, and htabkhiv@uncc.edu}}

\maketitle              % typeset the header of the contribution
\begin{abstract}
In recent years, we have seen a significant interest in data-driven deep learning approaches for video anomaly detection, where an algorithm must determine if specific frames of a video contain abnormal behaviors. However, video anomaly detection is particularly context-specific, and the availability of representative datasets heavily limits real-world accuracy. Additionally, the metrics currently reported by most state-of-the-art methods often do not reflect how well the model will perform in real-world scenarios. In this article, we present the Charlotte Anomaly Dataset (CHAD). CHAD is a high-resolution, multi-camera anomaly dataset in a commercial parking lot setting. In addition to frame-level anomaly labels, CHAD is the first anomaly dataset to include bounding box, identity, and pose annotations for each actor. This is especially beneficial for skeleton-based anomaly detection, which is useful for its lower computational demand in real-world settings. CHAD is also the first anomaly dataset to contain multiple views of the same scene. With four camera views and over 1.15 million frames, CHAD is the largest fully annotated anomaly detection dataset including person annotations, collected from continuous video streams from stationary cameras for smart video surveillance applications. To demonstrate the efficacy of CHAD for training and evaluation, we benchmark two state-of-the-art skeleton-based anomaly detection algorithms on CHAD and provide comprehensive analysis, including both quantitative results and qualitative examination. The dataset is available at \url{https://github.com/TeCSAR-UNCC/CHAD}.
\keywords{Anomaly Detection  \and Dataset \and Computer Vision \and Deep Learning.}
\end{abstract}
\section{Introduction}\label{sec:Intro}

Video anomaly detection, which requires understanding if a video contains anomalous behaviors, is a popular but challenging task in computer vision. In addition to substantial research interest, many real-world applications greatly benefit from being able to determine if such anomalous behaviors are present. Parking lot surveillance is one such application, where being able to determine the presence of an anomalous action (e.g. fighting, theft, fainting) is paramount.

Current state-of-the-art (SotA) deep learning solutions take one of two approaches. The first is an appearance-based method, where the algorithm works directly on video frames. The second is the skeleton-based methodology, in which algorithms rely on extracted human pose data to understand human behaviors. Both methods require large amounts of quality data. Anomaly detection is particularly context-specific, so training data must also be representative of both the environment and the context of the target application. This need is amplified for unsupervised approaches, which try to learn the normal behaviors of a specific context and need many example frames to do so.

There are currently only a limited number of datasets for video anomaly detection. These datasets, while seeing continual growth in the amount of data provided, also tend to fall short regarding the number of normal frames per context (i.e., per scene). Additionally, no current video anomaly dataset provides the detection, tracking, and pose information required by skeleton-based methods, leaving them to rely on external algorithms to generate this data. Since there is no standard for this, it is difficult to determine how much of an approach's error is due to the noise in this generated data or from the algorithm itself. This is further obfuscated by the inconsistency of the metrics used in reporting performance. Of the three main metrics for anomaly detection, discussed in \secref{sec:metric}, most SotA approaches only report one. However, all of them are necessary for a full understanding of an algorithm's performance, especially in the real-world.

In this paper, we present the Charlotte Anomaly Dataset (CHAD), a high-resolution, multi-camera anomaly detection dataset in a parking lot setting. CHAD is designed to address the most challenging issues facing current video anomaly detection datasets. The first video anomaly dataset with multiple camera views of a single scene, CHAD has over 1.15 million frames capturing the same context. With over 1 million normal frames, CHAD places itself as the premiere video anomaly dataset for unsupervised methods, providing human detection, tracking, and pose annotations. Thanks to these annotations, CHAD allows for a more accurate standard, positioning itself as the best-in-class dataset for skeleton-based anomaly detection.

We also propose a new standard in the benchmarking and evaluation of real-world video anomaly detection. Included is a detailed discussion on metrics, the benefits and disadvantages of each, and how the use of all three is needed to truly understand an algorithm's performance. To demonstrate the efficacy of CHAD, we train two SotA skeleton-based approaches, report both single camera and multi-camera performance, and compare to those methods trained on other datasets. Additionally, we perform cross-validation on CHAD, and the ShanghaiTech Campus Dataset \cite{liu2018ano_pred}, demonstrating CHAD's suitability for enabling generalization and revealing it to be more challenging than its peers.

In summary, this paper has the following contributions:
\begin{itemize}
    \item We introduce CHAD, a high resolution, multi-camera video anomaly detection dataset in a parking lot setting. With over 1.15 million frames of a single context and detection, tracking, and pose annotations, CHAD positions itself as the best-in-class dataset for both unsupervised and skeleton-based anomaly detection methods.
    \item We propose a new standard in real-world video anomaly detection benchmarking and evaluation. We provide a detailed discussion on the metrics used, including the insights they provide. 
    \item To validate the efficacy of CHAD, we train and evaluate two SotA skeleton-based models with our proposed methodology. We further perform cross-validation on CHAD and ShanghaiTech Campus\cite{liu2018ano_pred}, demonstrating that CHAD is robust enough for generalization while being more challenging.
\end{itemize}
\section{Related Work} \label{sec:related}

\subsubsection{Anomaly Detection Algorithms}
Appearance-based methods utilize appearance and motion features generated directly from pixel data for detecting anomalies \cite{8379443, ZHOU2016358, Sultani_2018_CVPR, Tian_2021_ICCV, NIPS2014_5ca3e9b1, 9023261, 8658774, Liu_2018_CVPR}. These methods generally achieve high accuracies in their context at the cost of high computation. Skeleton-based methods utilizes high-level, low-dimensional human pose skeletons \cite{Rodrigues_2020_WACV, LUO2021332, markovitz2020graph, Morais_2019_CVPR, li2022human}. These skeletons are informative in the context of human behavior while requiring far less computation than working with raw video data. They are more privacy preserving, and they remove demographic biases. As such, researchers have found significant success in skeleton-based anomaly detection. 

\subsubsection{Anomaly Detection Datasets}

The CUHK Avenue Dataset \cite{abnormal2013lu} consists of nearly 31K frames captured from a single camera. Abnormal objects, walking in the wrong direction, and sudden movements are examples of anomalous behaviors in this dataset.

The UCSD Anomaly Detection Dataset \cite{5539872} consists of 19K frames overlooking pedestrian walkways. UCSD has been categorized into two subsets, each one covering a different view. UCSD Ped1 sees pedestrian movement perpendicular to the camera, while UCSD Ped2 sees movement parallel to the camera. UCSD contains positional information for localizing anomalies.

The Subway dataset \cite{adam2008robust} consists of two surveillance videos, the subway entrance, and exit. With a combined total of 139 minutes of video, this dataset counts behaviors such as running, loitering, and walking in the opposite direction of the crowd as anomalous behaviors.

Street Scene \cite{9093457} is a single scene anomaly detection dataset captured from a bird's eye view of a two lane street. Compared to most other datasets, Street Scene is relatively large at over 200K frames. Street Scene also contains non-human anomalies, such as illegally parked cars, dogs on the sidewalk, and cars making u-turns. 

The ShanghaiTech Campus dataset \cite{liu2018ano_pred} contains 13 different scenes taken from a campus setting. With over 317K frames, ShanghaiTech is one of the largest and most popular anomaly detection datasets available. However, ShanghaiTech has relatively few frames per context.

IITB-Corridor \cite{Rodrigues_2020_WACV} was the largest single-stationary-camera anomaly detection dataset that existed before CHAD. It contains nearly 440K frames in a campus setting. Recorded in high-resolution 1080p, it is the only continuous video anomaly detection dataset with a resolution comparable to CHAD.

The ADOC dataset \cite{Pranav_2020_ACCV} is captured from a single high-resolution camera over 24 hours in a campus setting. ADOC consists of 260K frames and adopts an approach of considering any low-frequency behavior to be anomalous. Assuming only walking is normal, they consider all other behaviors as anomalous, even relatively commonplace activities like walking with a briefcase, having a conversation, or a bird flying through the air. While this categorization works for ADOC's context, it is inconsistent with how other datasets define anomalous behaviors.

Specifically for supervised anomaly detection, UBnormal \cite{Acsintoae_CVPR_2022} is comprised entirely of synthetically generated videos. With a total of 236,902 frames, UBnormal is moderately large compared to other anomaly datasets, though with 29 scenes the average number of frames per scene is fairly low.

The NOLA dataset \cite{doshi2022rethinking} is another new dataset. Collected over an entire week, NOLA contains over 1.4 million frames including both day and night scenes. In contrast to most other anomaly datasets, NOLA uses a single moving camera instead of stationary cameras. The rapid movement of the camera introduces a massive change of context, making the video anomaly detection more challenging. Due to the way annotations are presented in the dataset and the lack of clarifying documentation, it is impossible to ascertain what constitutes an anomaly in the context of NOLA. As such, it is difficult to determine the efficacy of this dataset for anomaly detection, and fair comparison to other datasets is not feasible.

UCF Crime \cite{Sultani_2018_CVPR} and X-D Violence \cite{wu2020not} collect video clips from many different sources in varying contexts, as opposed to continuous recordings. This allows them to be enormous by anomaly dataset standards but is so fundamentally different in problem formulation that it could be considered a different task altogether. XD-Violence provides both video and audio, making it unique among video anomaly datasets.

All of these datasets bring their own benefits and have helped advance the field of video anomaly detection. However, while they all have their own strengths, each of them also provides its own challenges when it comes to training networks for the real-world. Some datasets are too small, either in overall frames or frames per scene. Some of them have strict definitions of normal behaviors that would be undesirable in a real-world context. Some have to contend with domain shift, either from taking a large amalgamation of clips from entirely different contexts or from training with synthetic actors and moving to real persons and objects when used in a real-world context. And while many of these datasets provide multiple contexts, none of them provide different views of the same context, as would be fairly common in a surveillance setting. Further, none of these datasets provide the human detection, tracking, and pose annotations needed for skeleton-based anomaly detection. It is impossible for a single dataset to fit every possible scenario.
\section{Data Collection and Setup}\label{sec:Production}
\begin{figure}[!htbp]
\vspace*{-\baselineskip}

    \centering
    \begin{subfigure}{0.85\textwidth}
      \includegraphics[width=1\textwidth]{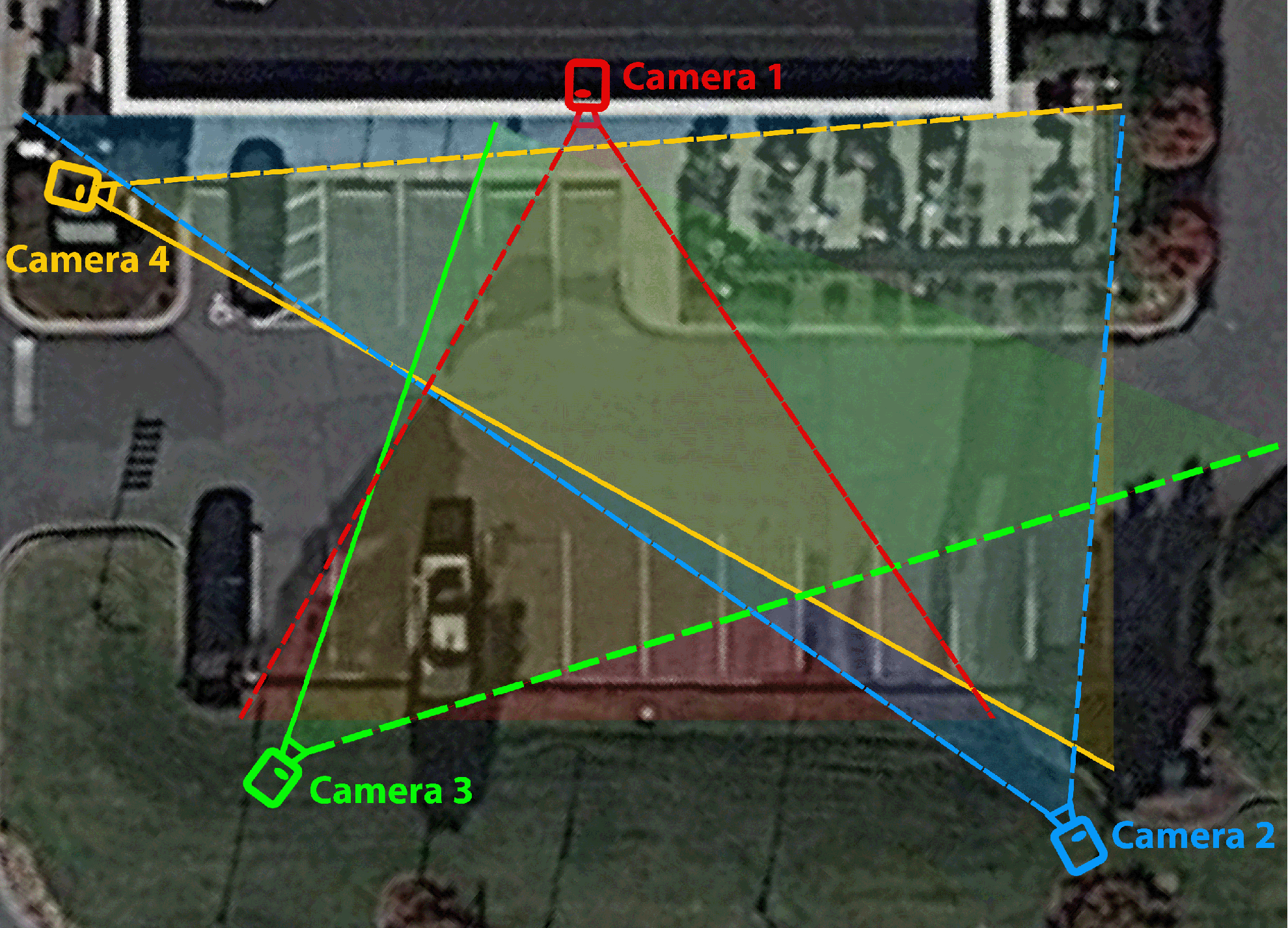}
      \caption{Birds-eye View}
      \label{fig:Ng1} 
    \end{subfigure}
    
    \begin{subfigure}{0.85\textwidth}
    \vspace{+5pt}
      \includegraphics[width=1\textwidth]{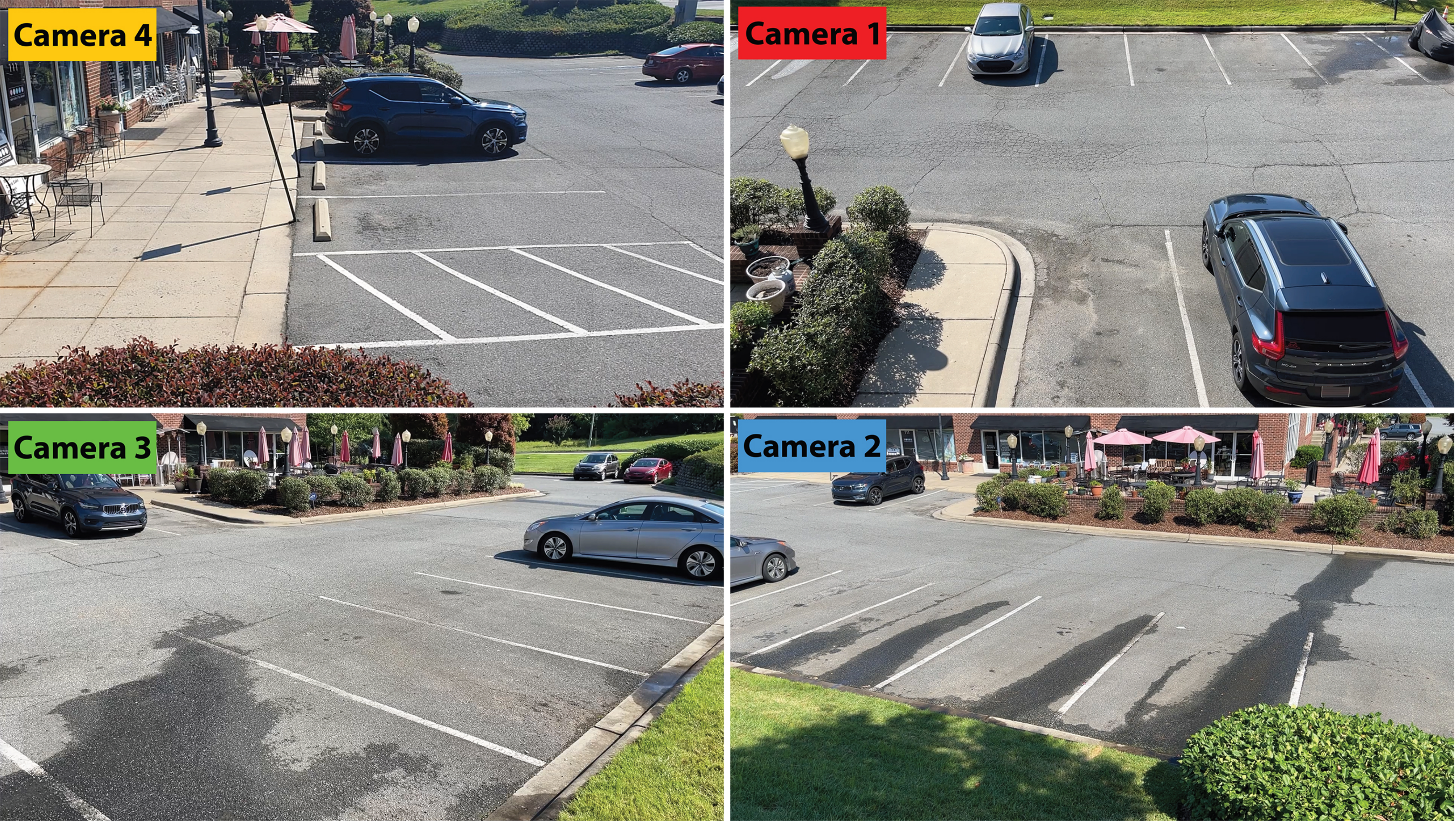}
      \caption{Camera View}
      \label{fig:Ng2}
    \end{subfigure}
    \vspace{-8pt}
    \caption{Approximate position and the views of the cameras.}
    \label{fig:view_points}
    \vspace*{-\baselineskip}
    \vspace*{-\baselineskip}
    \vspace*{-\baselineskip}
\end{figure}

Since anomaly detection is such a context-specific task, it is important that the data used to train algorithms is representative of their real-world environments. Often the disconnect between training data, and inference data leads to unsatisfying performance in the real-world \cite{alinezhad2022adg}. CHAD was designed to accurately mimic a real-world parking lot surveillance setting. The four cameras, as seen in \figref{fig:view_points}, were positioned to cover the same general scene, though their perspectives give them each a unique context compared to the others. Each video is recorded in full HD (1920x1080, 30fps), except camera 4 which is in standard HD (1280x720, 30fps), as seen in \tabref{tab:resolution}.

There are thirteen actors present in CHAD. The actors represent diverse demographics (gender, age, ethnicity, etc.) and each participates in both normal and anomalous clips. There are 22 classes of anomalous behaviors in CHAD, which can be seen in \tabref{tab:list}. This list has been curated in line with other state-of-the-art datasets \cite{liu2018ano_pred, abnormal2013lu, adam2008robust, 5539872}. All other actions present in CHAD (e.g. walking, waving, talking, etc.) are considered normal.

\begin{table}[!htbp]
\vspace*{-\baselineskip}
\vspace*{-\baselineskip}
    \centering
    \caption{Anomalous behaviors present in CHAD.}
    \label{tab:list}
    \renewcommand{\arraystretch}{1.2}
    \begin{tabular}{cccc}
        \multicolumn{4}{c}{\textbf{Type of Anomalous Behavior}} \\ \hline \hline
        \multicolumn{2}{c|}{Group Activities} & \multicolumn{2}{c}{Individual Activities} \\ \hline
        Fighting & \multicolumn{1}{c|}{Punching} & Throwing & Running \\
        Kicking & \multicolumn{1}{c|}{Pushing} & Riding & Falling \\
        Pulling & \multicolumn{1}{c|}{Slapping} & Littering & Jumping \\
        Strangling & \multicolumn{1}{c|}{Body Hitting} & Hopping & Sleeping \\
        Theft & \multicolumn{1}{c|}{Pick-Pocketing} &  &  \\
        Tripping & \multicolumn{1}{c|}{Playing with Ball} &  &  \\
        Chasing & \multicolumn{1}{c|}{Playing with Racket} &  & 
    \end{tabular}
    \vspace*{-\baselineskip}
    \vspace*{-\baselineskip}
\end{table}
\section{Annotation Methodology}\label{sec:Annotations}

CHAD contains four types of annotations: frame-level anomaly labels, person bounding boxes, person ID labels, and human keypoints.

\subsection{Anomaly Annotations}
We annotate anomalous behaviors at the frame level. This is, we mark the frame where the anomalous behavior begins, the frame where it ends, and every frame in between. This is done by hand, accounting for all the behaviors defined in \secref{sec:Production}. These frame-level labels are needed for both appearance-based and skeleton-based approaches. CHAD does not include anomaly localization labels.

\subsection{Person Annotations}
One of the innovations that sets CHAD above its peers is the inclusion of person annotations. In real-world scenarios, there is no access to hand-annotated data. The annotations must be generated through available tools and are not always perfect. We include generated person-annotations to ensure they are more representative of a real-world situation. This is by design, as a certain amount of noise is desirable in the dataset to assist models in learning how to deal with unclean data inherent in real-world situations \cite{chandra2019robusttp}. It also allows skeleton-based anomaly detection methods to have access to the processed data they need without having to spend time extracting it themselves. We hope this will make skeleton-based anomaly detection more accessible to researchers, leading to more innovation. It also sets a standard previously unavailable for how to generate this human detection, tracking, and pose information. With this standard, the variability based on the quality of input data is removed, leading to more precise and fair comparisons between approaches.

\subsubsection{Bounding Boxes}

The bounding box of a person refers to the upper and lower x and y coordinate limits they occupy in an image. Having quality bounding boxes for each individual and for every frame is doubly important for CHAD, as this localization is needed for the extraction of both person ID labels and human keypoints as well. For this reason, CHAD utilizes the popular object detection algorithm YOLOv4 \cite{bochkovskiy2020yolov4} for generating quality bounding boxes. Since CHAD is focused on anomalous human behavior, only the bounding boxes for people are used.

\subsubsection{Person ID Labels}
Anomaly detection algorithms often utilize temporal information to understand the behaviors of people. Particularly for skeleton-based methods, it is necessary to be able to associate the different poses of a person to that specific person across frames. Person ID labels provide this information, allowing for temporal tracking of individual persons in each video clip. Given the bounding box information generated previously, DeepSORT \cite{wojke2017simple} was utilized to provide tracking for persons through frames, generating unique person ID labels for each person in a video clip. For label stability, a three frame warm-up is used by DeepSORT before providing person ID labels. As such, the first two frames of each video clip are absent of person annotations. 

\subsubsection{Human Keypoints}

CHAD contains pose information in the form of human pose skeletons. These skeletons are made up of human keypoints, or points of interest on the human body. While there are several methods for defining what keypoints to use, CHAD follows the 17 keypoint methodology proposed by MS COCO \cite{lin2014microsoft}. Using the localization provided by the previously generated bounding boxes, keypoints are extracted using HRNet \cite{sun2019deep}, a prolific algorithm for human pose estimation used by many. To ensure we only provide quality keypoint annotations, we remove any person with low confidence (<50\%) for at least half of their keypoints (9+). While this leads to some frames where people are not detected, it helps reduce the overall noise of the data that is present.

\subsection{Annotation Smoothing}

The algorithms used to annotate CHAD are imperfect, and there are instances where people are completely missed at either the object detection or keypoint extraction stage. Combined with our purposeful removal of overly noisy data, this results in an undesirable number of missed persons. To compensate for this, we introduce annotation smoothing to CHAD, using high confidence annotations to help fill in the missing information. 

Given the relatively high frame rate of CHAD at 30 frames per second, it is a reasonable assumption that the positions and skeletons of a person will not drastically change between consecutive frames. As such, we can use linear interpolation to approximate the bounding box coordinates of each individual, assuming we have accurate detection at the start and end of the missing frames, and the number of missing frames is not too large. We choose 15 frames, or half a second, as a qualitative analysis showed this to be long enough to provide a significant benefit to annotation consistency, but not so long that the data it produced became unreliable. We apply the same smoothing technique to the keypoint annotations, with the same frame limitations. The details of smoothing are provided in the following equation:

\begin{equation}
    \mathbb{X}_i  = (\frac{\mathbb{X}_\mathcal{N} - \mathbb{X}_\mathcal{M}}{\mathcal{N}-\mathcal{M}})\times i + \mathbb{X}_\mathcal{M}
    \label{formaul:bbox}
\end{equation}

where $\mathbb{X}_i$ refers to a missing point (either bounding box or keypoint coordinate) at frame $i$, $\mathbb{X}_\mathcal{M}$ and $\mathbb{X}_\mathcal{N}$ refer to the two nearest matching points at frames $\mathcal{M}$ and $\mathcal{N}$ respectively, and where $\mathcal{M} < i < \mathcal{N}$ and $\mathcal{N}-\mathcal{M}+1 \leq 15$.

The added consistency in annotations created by this smoothing is particularly useful in the context of unsupervised learning. However, the confidence scores of keypoints generated by this smoothing are set to Null, so they can be easily discarded if undesired.

\begin{table}[]
\centering
\caption{Annotation availability in ShanghaiTech \cite{liu2018ano_pred}, CUHK \cite{abnormal2013lu}, UCSD \cite{5539872}, Subway \cite{adam2008robust}, IITB \cite{Rodrigues_2020_WACV}, Street Scene \cite{9093457}, UBnormal \cite{Acsintoae_CVPR_2022}, and CHAD (Ours).  \text{*} partially annotated, $-$ not annotated.}
\renewcommand{\arraystretch}{1.2}
\begin{tabular}{c||c|c|c|c|c}

\multirow{2}{*}{\textbf{Dataset}} & \multicolumn{2}{c|}{\textbf{Anomaly Annotations}}                                                                                                                                  & \multicolumn{3}{c}{\textbf{Person Annotations}}  \\ \cline{2-6}
                         & \multicolumn{1}{c|}{Frame-level} & \multicolumn{1}{c|}{Pixel-level} & \multicolumn{1}{c|}{Bounding Box} & \multicolumn{1}{c|}{ID Number} & \multicolumn{1}{c}{Keypoints}                    \\ \hline\hline

ShanghaiTech & \checkmark & \checkmark & $-$ & $-$ & $-$ \\ \hline
CUHK & \checkmark & \checkmark & $-$ & $-$ & $-$ \\ \hline
UCSD & \checkmark & \text{*} & $-$ & $-$ & $-$ \\ \hline
Subway & \checkmark &  \checkmark &  $-$ &  $-$ &   $-$ \\ \hline
IITB  & \checkmark &  \checkmark &  $-$ &  $-$ &   $-$ \\ \hline
Street Scene  & \checkmark &  \checkmark &  $-$ &  $-$ &   $-$ \\ \hline
UBnormal & \checkmark &  \checkmark &  \checkmark &  $-$ &   $-$ \\ \hline
\textbf{CHAD (Ours)} & \checkmark & $-$  & \checkmark & \checkmark & \checkmark 
\end{tabular}
\label{tab:annotations}
\vspace*{-\baselineskip}
% \vspace*{-\baselineskip}
\end{table}
\section{CHAD Statistics}\label{sec:Stats}

With over 1.15 million frames, CHAD is the largest anomaly detection dataset available that is recorded from continuous videos captured from stationary cameras, and includes person annotations. As shown in \tabref{tab:frames}, CHAD has more than 2$\times$ the number of frames as the next largest dataset, providing a substantial amount of learnable data. Additionally, CHAD has over 1 million frames of purely normal behaviors, which are required for unsupervised methods that rely on learning the normal to understand the anomalous. This is nearly 3$\times$ more than can be found in other datasets. The 59K anomalous frames in CHAD are comprised of the 22 anomalous behaviors presented in \tabref{tab:list}. To facilitate supervised, unsupervised, and semi-supervised approaches, CHAD includes two splits for training and testing. The \textit{unsupervised split} has a training set comprised only of normal behaviors, while the test set contains both normal and anomalous behaviors. The details of the \textit{unsupervised split} can be found in \tabref{tab:frames}. For the \textit{supervised split}, the normal and anomalous frames were distributed uniformly between the training and test sets, with 60\% of each belonging to the training set and 40\% to the test set.

More than just the amount of data, CHAD benefits from having high quality image data. As discussed in \secref{sec:Production}, CHAD was recorded from four high-resolution cameras with an overlapping view of a scene. Recorded at 30 FPS, CHAD not only boasts a higher resolution and frame rate than other datasets, shown in \tabref{tab:resolution}, but also presents data in a format representative of modern real-world surveillance systems. While resolution and frame rate are indicators of overall video quality and the amount of data present in each frame, they can not convey how much of that data is actually useful for learning. Difference of Gaussian \cite{crowley1984representation} is an image processing method that has been used to simulate how the human eye extracts visual details of an image for neural processing \cite{lv2015difference}. More simply, it creates a visual illustration of the density and richness of the features in an image. This allows us to visually analyze the quality of the data present in each dataset by comparing the Difference of Gaussian between them.

\begin{figure}[h]
\vspace*{-\baselineskip}
        \centering
               \includegraphics[width=1\linewidth]{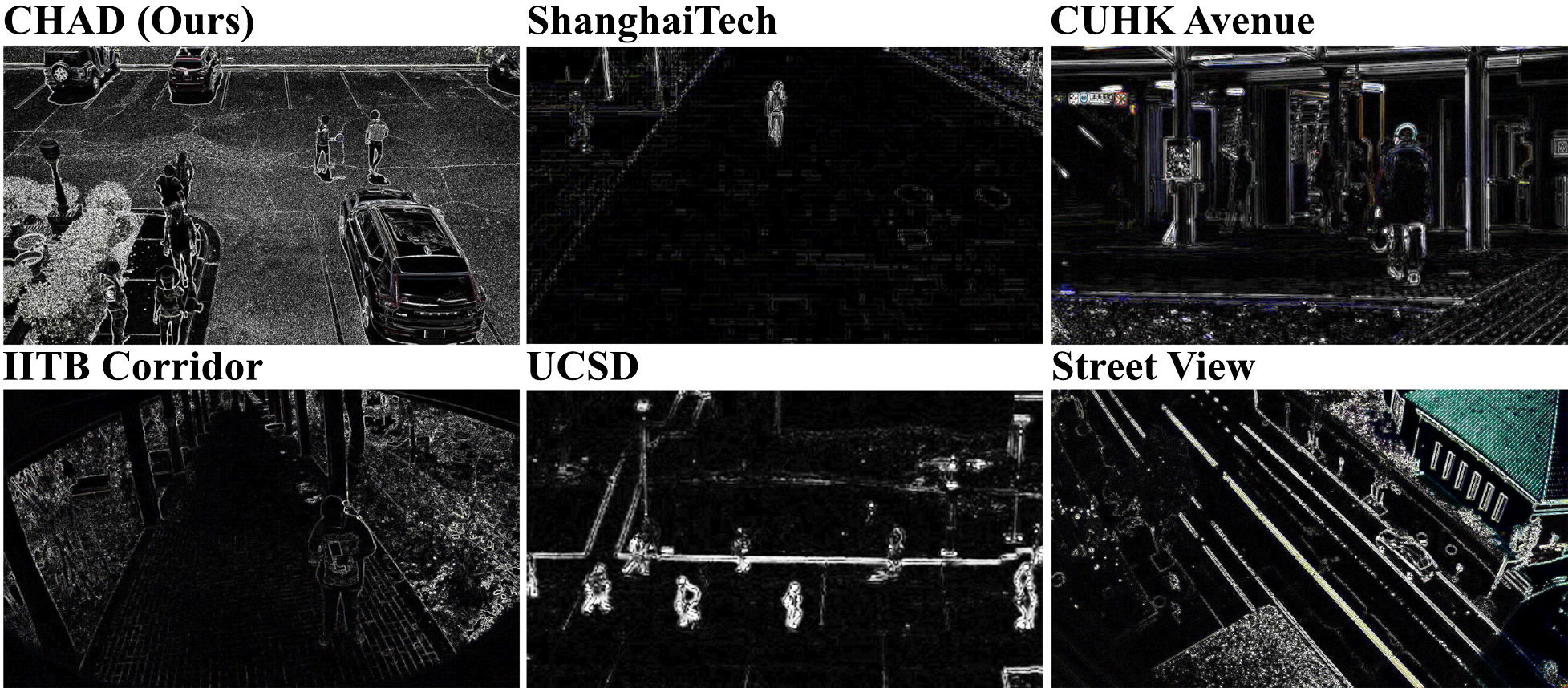}
                \caption{Visualization of Difference of Gaussian in Shanghai \cite{liu2018ano_pred}, CUHK \cite{abnormal2013lu}, UCSD \cite{5539872}, IITB \cite{Rodrigues_2020_WACV}, Street Scene \cite{9093457}, and CHAD (Ours). UCSD cropped to fit. All brightened for readability.}
                \label{fig:DoG}
% \vspace*{-\baselineskip}
% \vspace*{-\baselineskip}
% \vspace*{-\baselineskip}
\end{figure}

We visualize the Difference of Gaussian for a single frame of each dataset in \figref{fig:DoG}. We set a Gaussian blur radius of one pixel to maximize the precision of the resulting representation. Looking at the images, CHAD very clearly presents the most detail. This was anticipated due to its high resolution, but the amount by which it surpasses the other datasets far exceeded expectations. Fine details in the persons, clothing, vehicles, and the environment are clear, granting an accurate perception of the original image. IITB-Corridor \cite{Rodrigues_2020_WACV} is the only other dataset with 1080p images. However, the Difference of Gaussian tells a different story. While there are details present in the environment, they are comparably indistinct. Even in the brightened image, it is difficult to tell there is a person in the image. This demonstrates a surprising lack of rich features in the IITB-Corridor, despite the resolution. 

\begin{table}[!htbp]
\vspace*{-\baselineskip}
\centering
\caption{Resolution and frame rate in Shanghai \cite{liu2018ano_pred}, CUHK \cite{abnormal2013lu}, UCSD \cite{5539872}, Subway \cite{adam2008robust}, IITB \cite{Rodrigues_2020_WACV}, Street Scene \cite{9093457}, UBnormal \cite{Acsintoae_CVPR_2022}, and CHAD (Ours). N/A means Not Available.}
\renewcommand{\arraystretch}{1.2}
\begin{tabular}{cc||c|c}
\multicolumn{2}{c||}{\textbf{Dataset}} & \textbf{Resolution (Pixels)} & \textbf{Frame Rate (FPS)}\\ \hline \hline

\multicolumn{2}{c||}{Shanghai} & 856*480 & N/A\\ \hline
\multicolumn{2}{c||}{CUHK} & 640*360 & 25 \\ \hline
\multicolumn{1}{c|}{\multirow{2}{*}{UCSD}} & Ped1 & 238*158 & N/A \\ \cline{2-4}
\multicolumn{1}{c|}{} & Ped2 & 360*240 & N/A \\ \hline
\multicolumn{2}{c||}{Subway} & N/A & N/A\\ \hline
\multicolumn{2}{c||}{IITB} & 1920*1080 & 25 \\ \hline
\multicolumn{2}{c||}{Street Scene} & 1280*720 & 15\\ \hline
\multicolumn{2}{c||}{UBnormal} & varies & 30\\ \hline
\multicolumn{1}{c|}{\multirow{2}{*}{\textbf{CHAD (Ours)}}} & Scene 1-3 & 1920*1080 & 30 \\ \cline{2-4}
\multicolumn{1}{c|}{} & Scene 4 & 1280*720  & 30
\end{tabular}
\label{tab:resolution}
\vspace*{-\baselineskip}
\vspace*{-\baselineskip}
\end{table}

Street View, at the next highest resolution, shows much more detail and clarity than IITB-Corridor, though nowhere near the level of CHAD. What is most interesting is that while the building, car, and street boundaries are clear, it is difficult to notice the two people in the bottom left of the image. This is perhaps due to their relative size compared to the other objects mentioned and not necessarily indicative of a lack of features. Unsurprisingly, the lower resolution datasets, UCSD and CUHK Avenue, show sharp focal points (bright white pixels) but very little overall detail. Interestingly for ShanghaiTech \cite{liu2018ano_pred}, despite its slightly higher resolution, it presents a similar level of detail as Street Scene. However, due to the different camera perspectives, this translates into Shanghai providing better features for people, which is beneficial for its context.

Overall, we can see that CHAD not only has the best-in-class resolution and frame rate among anomaly detection datasets but also that the videos in CHAD are extremely feature rich, unrivaled among its peers. Additionally, there is a significant amount of background information irrelevant to person behaviors. The brightest spot in the Difference of Gaussian for CHAD is the foliage in the bottom left. This is noise - a distractor from information pertinent to anomaly detection. This means CHAD is not only more informative than other datasets but also suggests that it is more challenging as well. This level of challenge is needed if algorithms are to perform well in real-world scenarios, which are notorious for being more demanding than dataset benchmarks.

\begin{table}[!htbp]
\vspace*{-\baselineskip}
\centering
\caption{Dataset comparison for ShanghaiTech \cite{liu2018ano_pred}, CUHK \cite{abnormal2013lu}, UCSD \cite{5539872}, Subway \cite{adam2008robust}, IITB \cite{Rodrigues_2020_WACV}, Street Scene \cite{9093457}, UBnormal \cite{Acsintoae_CVPR_2022}, and CHAD (Ours). CHAD uses unsupervised split. N/A means Not Available.}
\begin{adjustbox}{max width=1\textwidth,center}
\begin{tabular}{c||ccccc|c|c}
\multirow{2}{*}{\textbf{Dataset}} & \multicolumn{5}{c|}{\textbf{Number of  Frames}}                                                                                                                                  & \multirow{2}{*}{\textbf{Scene(s)}} & \multirow{2}{*}{\textbf{Camera(s)}} \\ \cline{2-6}
                         & \multicolumn{1}{c|}{Total}  & \multicolumn{1}{c|}{Train} & \multicolumn{1}{c|}{Test} & \multicolumn{1}{c|}{Normal} & Anomalous &                                  &                         \\ \hline \hline

ShanghaiTech   &  \multicolumn{1}{c|}{317,398}         & \multicolumn{1}{c|}{274,515}         & \multicolumn{1}{c|}{42,883}           & \multicolumn{1}{c|}{300,308}       &      17,090     & \multicolumn{1}{c|}{13}            &              13           \\ \hline
CUHK             & \multicolumn{1}{c|}{30,652}       & \multicolumn{1}{c|}{15,328}         & \multicolumn{1}{c|}{15,324}           & \multicolumn{1}{c|}{26,832}       &     3,820      & \multicolumn{1}{c|}{1}            &                1         \\ \hline
UCSD                & \multicolumn{1}{c|}{18,560}       & \multicolumn{1}{c|}{9,050}         & \multicolumn{1}{c|}{9,210}           & \multicolumn{1}{c|}{12,919}       &     5,641      & \multicolumn{1}{c|}{2}            &              2           \\ \hline
Subway              & \multicolumn{1}{c|}{208,925}        & \multicolumn{1}{c|}{27,500}         & \multicolumn{1}{c|}{181,425}           & \multicolumn{1}{c|}{205,805}       &     3120      & \multicolumn{1}{c|}{2}            &              2         
\\ \hline
IITB             & \multicolumn{1}{c|}{483,566}         & \multicolumn{1}{c|}{301,999}         & \multicolumn{1}{c|}{181,567}           & \multicolumn{1}{c|}{375,288}       &     108,278     & \multicolumn{1}{c|}{1}            &              1     
\\ \hline

Street Scene              & \multicolumn{1}{c|}{203,257}             & \multicolumn{1}{c|}{56,847}         & \multicolumn{1}{c|}{146,410}           & \multicolumn{1}{c|}{N/A}       &    N/A     & \multicolumn{1}{c|}{1}            &              1     \\ \hline

UBnormal              & \multicolumn{1}{c|}{236,902}             & \multicolumn{1}{c|}{116,087}         & \multicolumn{1}{c|}{28,175}           & \multicolumn{1}{c|}{147,887}       &    89,015     & \multicolumn{1}{c|}{29}            &              -     \\ \hline

\textbf{CHAD (Ours)}              & \multicolumn{1}{c|}{1,152,649}          & \multicolumn{1}{c|}{1,026,174}         & \multicolumn{1}{c|}{126,475}           & \multicolumn{1}{c|}{1,093,477}       &    59,172       & \multicolumn{1}{c|}{1}            &                  4 
                   
\end{tabular}
\end{adjustbox}
\label{tab:frames}
\vspace*{-\baselineskip}
\end{table}
\section{Metrics and Measurements} \label{sec:metric}

There are three main metrics used for evaluating performance on anomaly detection datasets: Area Under the Receiver Operating Characteristic Curve, Area Under the Precision-Recall Curve, and Equal Error Rate. While none of these metrics are truly representative of overall performance, they each have their strengths and weaknesses, and, taken together, they can provide a comprehensive understanding of how an algorithm truly performs.

\subsection{Receiver Operating Characteristic Curve}

The Area Under the Receiver Operating Characteristic Curve (AUC-ROC) is simply the area under the curve when plotting the True Positive Rate (TRP) over the False Positive Rate (FPR) over various thresholds. This metric is specific to binary classification, such as determining if a video does or does not contain anomalous behavior. Generally, a higher AUC-ROC indicates that the model is better at separating inputs into their corresponding classes. The ROC curve itself also helps give insight into the trade-off between TPR and FPR at different thresholds \cite{fernandez2018learning}. However, AUC-ROC is not indicative of the final decisions of a model. The metric reports a final calculated number, and concluding useful information about the actual amount of False Negative Rate (FNR), when an anomaly is classified as normal is almost unfeasible. FNR is particularly important for real-world applications, and reporting it separately is crucial. Additionally, AUC-ROC is very sensitive to imbalances in data \cite{he2013imbalanced}, making it sub-optimal if one class is over represented, as is often the case with normal behaviors in anomaly datasets \cite{davis2006relationship}.

\subsection{Precision-Recall Curve}
Precision is the fraction of correct positive guesses over all positive guesses, while Recall is the fraction of correct positive guesses over all positive samples. The Precision-Recall Curve (PR) is useful for understanding how to balance Precision and Recall, while the area under this curve summarizes all the information represented in it. While AUC-PR heavily focuses on the positive class, it still accounts for the False Negative Rate (FNR) -- that is when the model classifies an anomaly as normal. As such, AUC-PR is a better metric for understanding the prediction ability of a model when compared to AUC-ROC \cite{saito2015precision}. Additionally, AUC-PR is better suited for highly imbalanced data \cite{saito2015precision}, making it better at evaluating the minority class \cite{he2013imbalanced}. As the minority class in anomaly detection usually refers to anomalous behaviors, this is an important quality for this context. However, AUC-PR is a final calculated number and it does not provide direct insight into the correct classification of negative samples, nor does it provide a measure for the number of incorrect decisions a model makes. Thus, much like AUC-ROC, AUC-PR provides an incomplete understanding of a model's performance.

\subsection{Equal Error Rate}
Another useful metric is the Equal Error Rate (EER) \cite{li2013anomaly}. Plotting the FNR and FPR over various thresholds produces two curves that intersect at one point. The value at the intersection is the EER and shows what threshold value allows the model to achieve a balance between FNR and FPR. In the context of video anomaly detection, the EER illustrates how many false alarms a model will raise and how many anomalous frames it will miss when at equilibrium. On its own, this metric offers little insight into the overall performance of a model \cite{Sultani_2018_CVPR}. However, when used as a complement to AUC-ROC and AUC-PR, a more complete understanding can be achieved.

\section{Evaluation}\label{sec:Evaluation}
All experiments were conducted on a server containing two Intel Xeon Silver 4114, one V100 GPU, and 256 GB of RAM. We performed each experiment (training and testing) five times, averaging the results to remove any potential skew due to variability. For each model, training is performed exactly as described in their respective papers unless otherwise specified. 

\subsection{Standard Validation}
To demonstrate CHAD's viability as an anomaly detection dataset, we train and evaluate two state-of-the-art skeleton-based models using the \textit{unsupervised split}. We select Graph Embedded Pose Clustering (GEPC) \cite{markovitz2020graph} and Message-Passing Encoder-Decoder Recurrent Neural Network (MPED-RNN) \cite{Morais_2019_CVPR} for their high accuracy and model availability. GEPC utilizes a spatio-temporal graph autoencoder, while MPED-RNN uses a two-headed structure with reconstruction and prediction.

Both models were trained on each of CHAD's four camera views individually, the results reported in \tabref{tab:results}. The most obvious observation is that both models were able to learn on CHAD. GEPC achieved an average AUC-ROC of 0.663 and AUC-PR of 0.619, while MPED-RNN achieved an average AUC-ROC of 0.718 and AUC-PR of 0.635. For both models, the AUC-ROC is noticeably higher than the AUC-PR. This is largely due to the overwhelming majority of normal frames in the data, which if properly classified will a significant boost to the AUC-ROC. AUC-PR, on the other hand, does not count True Negatives, and as such gives a more measured result for the imbalanced data. Additionally, GEPC achieved an EER of 0.378 and MPED-RNN an EER of 0.339. This means that, given the threshold at equilibrium, both models can expect to see between 34\% and 38\% of both normal frames and anomalous frames to be misclassified. This is important to understand when targeting real-world applications, where misclassification rates are more important than class separability.
\begin{table}[!htbp]
\vspace*{-\baselineskip}
\vspace*{-\baselineskip}
\caption{Evaluation of GEPC \cite{markovitz2020graph} and MPED-RNN \cite{Morais_2019_CVPR} on CHAD (Ours).}
\label{tab:results}
\renewcommand{\arraystretch}{1.2}
\centering
%\begin{adjustbox}{max width=1\columnwidth,center}
\begin{tabular}{c||c|c|c|c}
    \textbf{Model} & \textbf{Camera} & \textbf{AUC-ROC} & \textbf{AUC-PR} & \textbf{EER} \\ \hline \hline
    \multirow{4}{*}{GEPC}    & 1 & 0.673 & 0.636 & 0.363 \\ \cline{2-5} 
                             & 2 & 0.660 & 0.566 & 0.382 \\ \cline{2-5}
                             & 3 & 0.661 & 0.586 & 0.384 \\ \cline{2-5} 
                             & 4 & 0.656 & 0.688 & 0.382 \\ \hline \hline
                             & 1 & 0.747 & 0.715 & 0.303 \\ \cline{2-5} 
                MPED-RNN     & 2 & 0.691 & 0.567 & 0.349 \\ \cline{2-5}
                             & 3 & 0.771 & 0.584 & 0.331 \\ \cline{2-5} 
                             & 4 & 0.662 & 0.674 & 0.372
\end{tabular}
%\end{adjustbox}
\vspace*{-\baselineskip}
\vspace*{-\baselineskip}
\end{table}

\subsection{Cross Validation}
To illustrate CHAD's ability to train models that can generalize, we perform cross validation experiments with another anomaly dataset in the same domain. We choose the popular ShanghaiTech Campus Dataset \cite{liu2018ano_pred} for its relatively large size, its similar context to CHAD, and its proven track record in anomaly detection research. For these experiments, we use GEPC, as its multi-camera training methodology allows for a simple conversion to cross validation. For both CHAD and ShanghaiTech, a single model is trained for all cameras in one dataset, then tested on both datasets. The results can be seen in \tabref{tab:cross_val}.

\begin{table}[!htbp]
\vspace*{-\baselineskip}
\vspace*{-\baselineskip}
\caption{Cross-validation of GEPC \cite{markovitz2020graph} on ShanghaiTech \cite{liu2018ano_pred} and CHAD (Ours).}
\label{tab:cross_val}
\renewcommand{\arraystretch}{1.2}
\centering
%\begin{adjustbox}{max width=1\columnwidth,center}
\begin{tabular}{c||c|c|c|c|c}
\textbf{Model} & \textbf{Train} & \textbf{Test} & \textbf{AUC-ROC} & \textbf{AUC-PR} & \textbf{EER} \\ \hline \hline
\multirow{4}{*}{GEPC} & \multirow{2}{*}{CHAD (Ours)} & CHAD (Ours) & 0.649 & 0.587 & 0.385 \\ \cline{3-6} 
 &  & ShanghaiTech & 0.728 & 0.637 & 0.326 \\ \cline{2-6} 
 & \multirow{2}{*}{ShanghaiTech} & CHAD (Ours) & 0.639 & 0.572 & 0.399 \\ \cline{3-6} 
 &  & ShanghaiTech & 0.741 & 0.657 & 0.315
\end{tabular}
\vspace*{-\baselineskip}
%\end{adjustbox}
\end{table}

The first thing to notice is that models trained on CHAD perform well on ShanghaiTech, and models trained on ShanghaiTech perform well on CHAD. This is logical, as the contexts for the two datasets (i.e. setting, camera views, anomalous behaviors) are quite similar. In all metrics, the validation of models across datasets performs within 1-2\% of models validated on their parent datasets, showing that models trained on either can generalize quite well given their similar contexts.

Another trend seen in \tabref{tab:cross_val} is that for all metrics, models tend to achieve lower scores (or higher in the case of EER) on CHAD than they do on ShanghaiTech. Since both models performed equally well in cross validation, the logical assumption is that CHAD's test set is more challenging than ShanghaiTech's. This is in part due to the additional noise and distractors present in CHAD, as explained in \secref{sec:Stats}. The other major factor is the inclusion of very subtle and complex anomalies in CHAD. Pick-pocketing is subtle by design, as most pick-pockets are trying not to be seen. Littering is also quite complex to learn, especially for a model that relies solely on human keypoints. Combined with the sheer size of CHAD's test set (3$\times$ that of ShanghaiTech's), this makes for a very challenging dataset for current anomaly detection algorithms.
\section{Conclusion}\label{sec:Conclusion}

This paper presented the Charlotte Anomaly Dataset (CHAD). Consisting of more than $1.15m$ high-resolution frames of a single scene, CHAD is the largest available anomaly detection dataset consisting of continuous video from stationary cameras. In addition to frame-level anomaly labels, CHAD goes further than other datasets and provides bounding-box, person ID, and human keypoints annotations, enabling a unified benchmarking standard for both skeleton and appearance-based anomaly detection. Additionally, this paper assesses three metrics for anomaly detection and proposes their use in combination as a new standard for real-world video anomaly detection.

\subsubsection{Acknowledgements} 
This research is supported by the National Science Foundation (NSF) under Award No. 1831795 and NSF Graduate Research Fellowship Award No. 1848727. In addition, this research is IRB approved under Document Number IRBIS-17-0307.

\bibliographystyle{splncs04}
\bibliography{references}
\end{document}